\RequirePackage[T1]{fontenc}
\documentclass[letterpaper,10pt,conference]{ieeeconf}
\IEEEoverridecommandlockouts
\usepackage{cite}
\usepackage[hyphens]{url}
\usepackage{xcolor}
\definecolor{accvblue}{rgb}{0,0,1}
\usepackage{accvabbrv}
\usepackage{graphicx}
\usepackage{booktabs}
\usepackage{amsmath}
\usepackage{amssymb}
\usepackage{float}

\usepackage[hidelinks]{hyperref}
\usepackage{orcidlink}
\usepackage{doi}
\usepackage{xurl}
\usepackage{flushend}
\hypersetup{
  pdftitle={SPACE-CLIPv2: Decoding Local Geometry from Frozen CLIP for Monocular Depth Estimation},
  pdfauthor={Hyun Song, Taewan Cho, Kangmin Kim, Andrew Jaeyong Choi},
  pdfsubject={Monocular depth estimation with frozen vision-language representations},
  pdfkeywords={SPACE-CLIPv2, CLIP, monocular depth estimation},
  pdflang={en-US}
}

\begin{document}

\title{\LARGE\bf SPACE-CLIPv2: Decoding Local Geometry from Frozen CLIP for Monocular Depth Estimation}

% Shared affiliation placed directly below the author names.
\author{Hyun Song, Taewan Cho, Kangmin Kim, and Andrew Jaeyong Choi$^{*}$\\[0.8ex]
{\normalsize School of Computing, iRASC Lab., Gachon University}\\[0.3ex]
{\small\texttt{\{songhyun, taewan2002, kgmin2413, andrewjchoi\}@gachon.ac.kr}}%
\thanks{\protect\raggedright $^{*}$Corresponding author: Andrew Jaeyong Choi.}%
}

\maketitle
\raggedbottom
\setlength{\emergencystretch}{1em}

\begin{abstract}
Vision-language foundation models such as CLIP provide strong semantic
representations, but their patch tokens are not directly optimized for dense
metric geometry. SPACE-CLIP showed that frozen CLIP features can support
monocular depth estimation through layer-group feature fusion, yet it leaves
open how neighboring CLIP tokens should be combined to recover fine local
structure. We present SPACE-CLIPv2, a frozen-backbone depth decoder that
aggregates fixed local neighborhoods in CLIP token space. At selected decoder
stages, the model samples a fixed token stencil, predicts aggregation weights,
and injects the resulting response through a gated residual update. A
token-space high-pass branch further preserves shallow local contrast. On NYU
Depth V2, SPACE-CLIPv2 improves over a matched SPACE-CLIP baseline, while
five-seed experiments consistently favor fixed over learned-offset sampling.
Zero-shot iBims-1 evaluation further improves boundary and planar-geometry
measures. These results support constrained local token aggregation as a
practical mechanism for decoding geometry from frozen CLIP representations.
\end{abstract}

\section{Introduction}
\label{sec:intro}

\indent Monocular depth estimation recovers dense 3D structure from a single RGB image.
Unlike image recognition, it depends strongly on local spatial cues such as
object boundaries, depth discontinuities, and surface transitions. This makes
frozen foundation encoders useful but difficult to adapt: their representations
often encode rich scene information, yet they are not trained to produce dense
metric geometry. Vision-language models such as CLIP~\cite{radford2021learning}
have therefore been explored for depth prediction through language-guided or
foundation-model adaptation strategies~\cite{zhang2022depthclip,chen2024depthblip2,zavadski2024primedepth}.
These works show that pretrained features can support geometric prediction, but
the decoder must still recover local structure from patch-token features learned
for recognition or image--text alignment.

\begin{figure*}[t]
  \centering
  \includegraphics[width=0.80\textwidth]{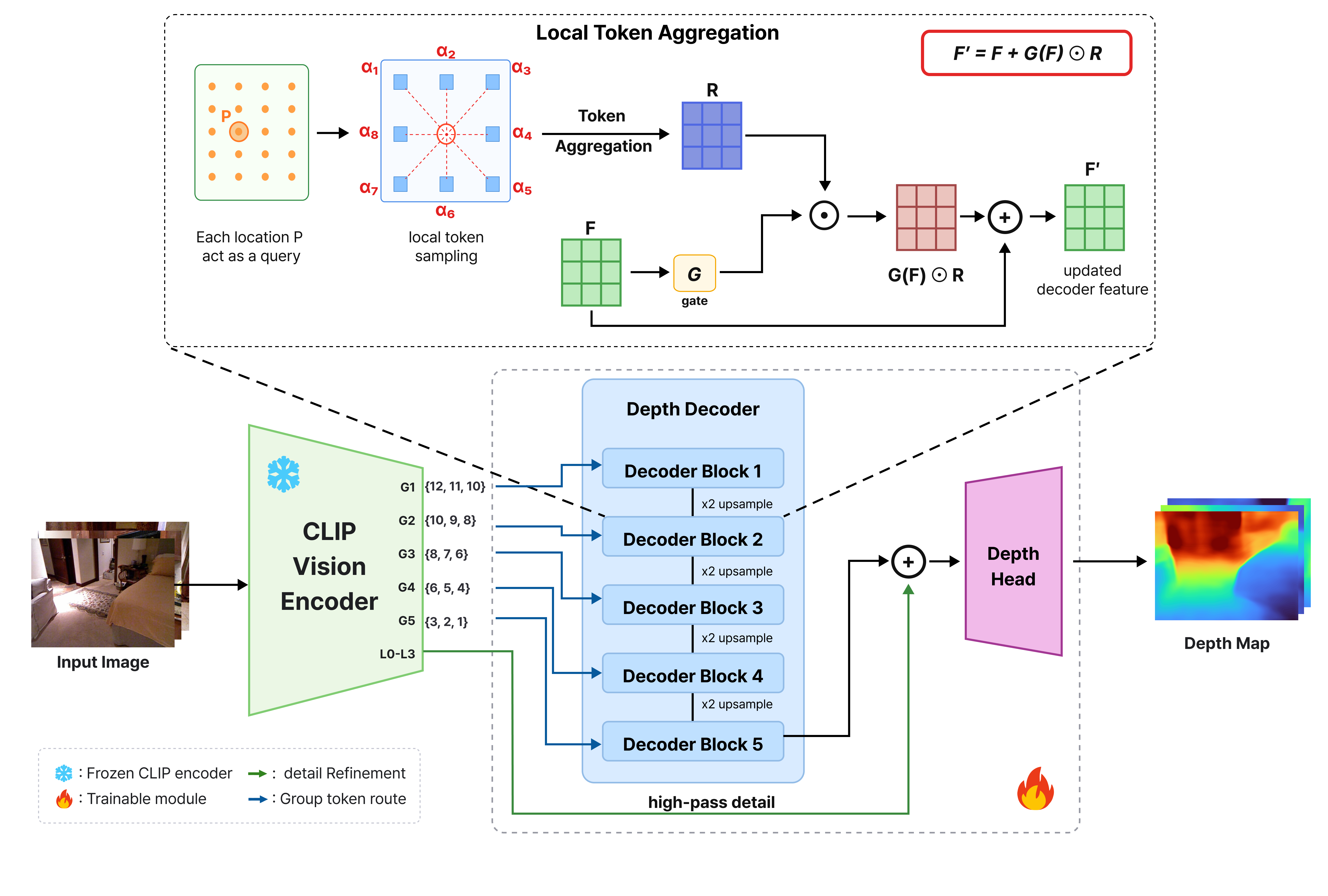}
  \caption{\textbf{SPACE-CLIPv2 overview.}
  A frozen CLIP vision encoder provides grouped patch-token maps to a trainable
  depth decoder. SPACE-CLIPv2 augments layer-level feature fusion with fixed
  local token aggregation: each selected decoder stage reads a fixed token-space
  neighborhood, predicts aggregation weights, and injects the aggregated
  response through a gated residual update. The shallow token-detail branch is
  included as a secondary refinement path and is held fixed in the core
  ablation.}
  \label{fig:method_overview}
\end{figure*}

SPACE-CLIP~\cite{cho2026spaceclipspatialperceptionadaptive} shows that frozen
CLIP visual features can be used for monocular depth estimation without updating
the CLIP backbone or using text prompts at inference time. Its decoder adapts
multi-layer CLIP features through layer-level feature selection and fusion,
indicating that CLIP preserves depth-relevant structural and semantic cues.
However, layer-level fusion does not explicitly specify how the decoder should
combine nearby CLIP tokens. For dense depth prediction, this local interaction
is important because many errors arise from nearby-patch relationships around
boundaries, thin objects, and changing surface orientations.

SPACE-CLIPv2 aggregates nearby CLIP tokens before dense prediction. At selected
decoder stages, the model reads a fixed token-space neighborhood around each
location and aggregates the sampled CLIP responses. The resulting response is
injected through a gated residual update, allowing the decoder to control how
strongly neighborhood evidence modifies its feature map while keeping the CLIP
encoder frozen. Instead of learning deformable sampling offsets from the final
depth loss, we use a fixed local stencil and learn the aggregation and residual
weighting. This gives the decoder a simple locality prior without adding a
trainable offset prediction branch.

We also study a token-space refinement branch that uses early CLIP token maps to
preserve local contrast. Rather than adding a separate RGB structural branch,
this refinement operates directly in token space and injects detail-oriented
features near the end of the decoder.

We evaluate SPACE-CLIPv2 on NYU Depth V2~\cite{silberman2012indoor} under a
matched frozen-backbone and text-free inference setting. Local aggregation
improves over both the reproduced SPACE-CLIP baseline and a decoder without
local aggregation. Across five randomly selected seeds, fixed-local sampling
consistently outperforms learned-offset sampling. We further evaluate the
NYU-trained checkpoints zero-shot on iBims-1~\cite{koch2019evaluation} and
DIODE~\cite{vasiljevic2019diode}. The iBims-1 results show substantial
improvements in boundary errors together with improved planar-geometry
measures, while indoor DIODE provides complementary cross-domain evidence.
Together, these results support fixed local aggregation as a simple and robust
decoder-side locality prior for frozen CLIP features.

Our contributions are summarized as follows:
\begin{itemize}
    \item We propose a fixed local token aggregation module that samples
    neighboring CLIP tokens and combines them with learned aggregation weights
    for dense depth decoding.
    \item We introduce a gated residual update that injects aggregated token
    responses into the depth decoder while keeping the CLIP image encoder
    frozen.
    \item We design and validate a token-space high-pass refinement branch that
    uses early CLIP token maps as a complementary detail path for preserving
    local contrast during depth decoding.
\end{itemize}

\section{Related Work}
\label{sec:related_work}

\subsection{Monocular Depth Estimation}

Monocular depth estimation predicts dense scene geometry from a single RGB image.
Benchmarks such as NYU Depth V2~\cite{silberman2012indoor} have driven progress
in indoor depth estimation by providing paired RGB and metric depth supervision.
Early deep models combined global scene prediction with local refinement for
single-image depth estimation~\cite{eigen2014depth}. Recent depth estimators
commonly rely on strong visual backbones, adaptive depth-bin formulations, and
multi-scale decoder designs to recover both global scene layout and local object
boundaries~\cite{ranftl2021vision,bhat2021adabins,li2024binsformer}. Larger
metric-depth and foundation-style systems further improve generalization by
combining relative and metric depth, mixed camera models, or large-scale
unlabeled data~\cite{bhat2023zoedepth,yin2023metric3d,yang2024depthanything}.
Other recent approaches use robust distillation or generative priors for
monocular depth estimation~\cite{chen2024structurecentric,zavadski2024primedepth}.
However, many of these approaches train or fine-tune depth-specific encoders.
Our work studies a stricter setting in which the visual encoder remains frozen
and all adaptation is pushed to the decoder.

\subsection{Frozen CLIP Representations for Dense Prediction}

CLIP~\cite{radford2021learning} learns transferable visual representations from
large-scale image--text supervision on top of patch-based visual encoders
popularized by Vision Transformers~\cite{dosovitskiy2021image}. Prior work on
CLIP-based dense prediction shows that image--text pre-training can be adapted
beyond image-level recognition~\cite{rao2022denseclip}. However, CLIP is trained
with image-level contrastive objectives rather than dense geometric supervision,
and its patch-token maps are coarse and semantically biased. Monocular depth
estimation, in contrast, requires fine spatial detail and metric structure. This
gap makes decoder design especially important when the CLIP backbone is kept
frozen.

\subsection{CLIP-Based Monocular Depth Estimation}

Several recent methods study how CLIP can be used for monocular depth
estimation. DepthCLIP~\cite{zhang2022depthclip} shows that CLIP representations
can provide useful depth cues through semantic alignment. Follow-up methods
improve CLIP-based depth estimation through prompt learning, task-specific
adaptation, or stronger decoder designs~\cite{hu2024learningclipdepth,auty2023learningpromptclipdepth}.
Related language-guided VLM approaches such as
DepthBLIP-2~\cite{chen2024depthblip2} also suggest that linguistic context can
support depth reasoning, although our setting remains text-free at inference time.
PureCLIP-Depth~\cite{miya2026pureclipdepth} further investigates depth prediction
under a frozen-CLIP setting, while CaBins~\cite{son2024cabins} and
CLIP2Depth~\cite{kim2024clip2depth} report competitive CLIP-based monocular
depth results.

SPACE-CLIP~\cite{cho2026spaceclipspatialperceptionadaptive} is the direct
predecessor to our work. It demonstrates that frozen CLIP features can support
monocular depth estimation under text-free inference and frozen-backbone
constraints. Its decoder mainly adapts CLIP through layer-group feature fusion,
such as combining structural and semantic layer groups. SPACE-CLIPv2 keeps the
same frozen-backbone and text-free setting, but shifts the focus from deciding
\emph{which layers} to fuse to deciding \emph{how local token neighborhoods}
should be decoded once those layers are available.

\subsection{Local Geometry Decoding from Frozen Tokens}

Dense depth prediction depends on local evidence as well as global scene
structure. Useful cues include object boundaries, surface discontinuities, and
fine depth transitions. A natural strategy is therefore to aggregate nearby
token responses. Local windows already serve as a standard locality prior in
visual transformers and dense prediction backbones~\cite{liu2021swin}. However,
under a frozen backbone, unconstrained offset learning may be difficult because
the feature space cannot adapt to the learned sampling locations. This motivates
a simpler design based on fixed local sampling.

SPACE-CLIPv2 follows this direction. It reads from a fixed neighborhood in CLIP
token space, learns aggregation weights for the sampled tokens, and injects the
result through a gated residual update. We also analyze a lightweight
token-detail branch near the depth head to test whether shallow CLIP tokens
provide complementary high-frequency contrast.

\section{Method}
\label{sec:method}

Fig.~\ref{fig:method_overview} illustrates SPACE-CLIPv2. Given a single RGB
image, the model extracts patch-token representations from a frozen CLIP vision
encoder~\cite{radford2021learning}. SPACE-CLIPv2 extends
the text-free depth decoder of
SPACE-CLIP~\cite{cho2026spaceclipspatialperceptionadaptive}. A depth decoder
then progressively upsamples its feature maps while receiving grouped CLIP
patch-token maps at each decoder stage. The main decoder
update is augmented by fixed local token aggregation, which samples a fixed
local stencil from the CLIP token grid and injects the aggregated response
through a gated residual path. We additionally study a token-detail refinement
branch that uses shallow CLIP token maps to preserve local contrast near the
depth prediction head.

\subsection{Frozen CLIP Token Extraction}

Let $I$ denote an input RGB image. We process $I$ with a frozen CLIP vision
encoder $E_{\mathrm{CLIP}}$ and extract hidden token sequences from selected
transformer layers:
\begin{equation}
    \{Z^\ell\}_{\ell=0}^{L} = E_{\mathrm{CLIP}}(I).
\end{equation}
For each selected layer $\ell$, non-patch tokens are removed and the remaining
patch tokens are reshaped into a spatial token map
$T^\ell \in \mathbb{R}^{h \times w \times c}$. These CLIP token maps provide the
frozen foundation-model features used by the adaptation modules. The CLIP text
encoder and text prompts are not used. The CLIP vision encoder remains frozen,
and all trainable adaptation is placed in the depth decoder.

\subsection{Layer-Grouped Decoder Routing}

Each decoder block receives a stage-specific group of adjacent CLIP layers. For
decoder stage $s$, we define a layer group $\mathcal{G}_s$ and form a grouped
token map by averaging the corresponding patch-token maps:
\begin{equation}
    \bar{T}_s =
    \frac{1}{|\mathcal{G}_s|}
    \sum_{\ell \in \mathcal{G}_s} T^\ell .
\end{equation}
In our implementation, the grouped layers follow a coarse-to-fine routing:
\begin{equation}
\begin{aligned}
    \mathcal{G}_1 &= \{12,11,10\}, &
    \mathcal{G}_2 &= \{10,9,8\}, \\
    \mathcal{G}_3 &= \{8,7,6\}, &
    \mathcal{G}_4 &= \{6,5,4\}, \\
    \mathcal{G}_5 &= \{3,2,1\}.
\end{aligned}
\end{equation}
The partially overlapping groups are designed to preserve feature continuity
across neighboring decoder stages while maintaining a coarse-to-fine routing
over the frozen CLIP hierarchy.
The grouped token maps are projected to the channel width of the corresponding
decoder block. The decoder proceeds from deep to shallow features, applying
upsampling between blocks to recover spatial resolution.

\subsection{Fixed Local Token Aggregation}

For a decoder feature map $F_s$, each spatial location acts as a dense query
that reads local evidence from the corresponding frozen CLIP token map
$\bar{T}_s$. Let $\Omega=\{\delta_k\}_{k=1}^{K}$ denote a fixed local stencil.
The decoder samples this fixed set of neighboring token locations and learns how
to weight their responses. Unlike deformable sampling
methods~\cite{dai2017deformable,zhu2021deformabledetr}, our sampling
coordinates remain fixed, and only the aggregation weights are predicted from
the decoder feature. A projected token value map $V_s$ is sampled by bilinear
interpolation:
\begin{equation}
\begin{aligned}
  \tilde{R}_s(p) &= \sum_{k=1}^{K}\alpha_{s,k}(p)\,
  \mathcal{B}\!\left(V_s,p+\lambda\delta_k\right),\\
  \alpha_{s,k}(p) &= \frac{\exp(a_{s,k}(p))}
  {\sum_{j=1}^{K}\exp(a_{s,j}(p))}.
\end{aligned}
\end{equation}
Here, $\mathcal{B}(\cdot)$ denotes bilinear sampling on the token grid,
$\lambda$ is a fixed sampling radius, and $a_{s,k}(p)$ is the decoder-predicted
score for the $k$-th fixed sample. When token saliency is enabled, $V_s$ denotes
the saliency-modulated token value map.

Because the aggregation weights are normalized with a softmax, the pre-mixing
response $\tilde{R}_s(p)$ is a convex combination of sampled token values:
\begin{equation}
\begin{gathered}
  \alpha_{s,k}(p)\geq0,\qquad \sum_{k=1}^{K}\alpha_{s,k}(p)=1,\\
  \tilde{R}_s(p)\in\mathrm{conv}
  \left\{\mathcal{B}\!\left(V_s,p+\lambda\delta_k\right)\right\}_{k=1}^{K}.
\end{gathered}
\end{equation}

\subsection{Gated Residual Injection}

The aggregated local token response is not directly added to the decoder
feature. It is first passed through a lightweight mixing block and then injected
through a gated residual update:
\begin{equation}
    R_s = M_s(\tilde{R}_s),
    \qquad
    F'_s = F_s + \gamma_s\,G_s(F_s)\odot R_s .
\end{equation}
Here, $M_s$ denotes the local mixing block, $G_s(F_s)$ is a decoder-derived
sigmoid gate, $\gamma_s$ is a learned residual scale, and $\odot$ denotes
element-wise modulation. Since $G_s(F_s)\in[0,1]$, the update magnitude is
controlled by both the gate and the residual scale:
\begin{equation}
    \|F'_s - F_s\|_2
    =
    \|\gamma_s G_s(F_s)\odot R_s\|_2
    \leq
    |\gamma_s|\,\|R_s\|_2 .
\end{equation}
This update combines an identity-preserving residual path with learned sigmoid
gating~\cite{he2016deep,srivastava2015highway}.
This residual formulation lets the decoder decide how strongly aggregated CLIP
token responses should affect the depth feature while preserving the original
decoder representation through the identity path. In practice, the local
aggregation residual is initialized with a small scale and a negative gate bias,
so the module starts close to the original decoder before learning local
corrections.

\subsection{Token-Space Detail Refinement}

Motivated by the shallow structural pathway of
SPACE-CLIP~\cite{cho2026spaceclipspatialperceptionadaptive}, we further
evaluate whether shallow CLIP token maps can provide complementary local
contrast near the final depth prediction head. This branch operates on CLIP
token maps rather than on the raw RGB image. For a selected shallow token map
$T$, we define its local high-pass component as
\begin{equation}
    H(T) = T - \mathrm{AvgPool}_{3\times3}(T).
\end{equation}
We evaluate raw token detail $T$ and raw-plus-high-pass detail. Adding the
high-pass residual back to the original token map with unit gain corresponds
to the token-space analogue of classical unsharp
masking~\cite{polesel2000unsharp}:
\begin{equation}
    T_{\mathrm{raw+hp}} = T + H(T)
    = 2T - \mathrm{AvgPool}_{3\times3}(T).
\end{equation}
The resulting shallow token detail features are projected, averaged across the
selected shallow layers, passed through a lightweight mixing block, and fused
with the final decoder feature through a gated residual update. This branch is
used to test whether early CLIP token maps provide local contrast that is
complementary to the main local aggregation path.

\subsection{Depth Prediction and Loss}

After the final decoder block and optional token-detail fusion, a lightweight
depth head maps the final decoder feature to a dense depth prediction:
\begin{equation}
    \hat{D} = g(F_{\mathrm{final}}).
\end{equation}
For the reported main experiments, the model is supervised on valid depth pixels
with a weighted combination of scale-invariant logarithmic
loss~\cite{eigen2014depth,bhat2021adabins} and structural similarity
loss~\cite{wang2004image}.
Let $D$ be the ground-truth depth map and $\mathcal{V}$ the valid-pixel set:
\begin{equation}
    \mathcal{L}
    =
    (1-\eta)\mathcal{L}_{\mathrm{SILog}}(\hat{D},D;\mathcal{V})
    +
    \eta\mathcal{L}_{\mathrm{SSIM}}(\hat{D},D;\mathcal{V}).
\end{equation}
The SILog term is computed in log-depth space:
\begin{equation}
\begin{aligned}
  e_i &= \log\hat{d}_i-\log d_i,\quad i\in\mathcal{V},\\
  \mathcal{L}_{\mathrm{SILog}} &= 10\sqrt{\mathrm{Var}(e)+(1-\lambda)\bar{e}^{\,2}}.
\end{aligned}
\end{equation}
Only the decoder and adaptation modules are updated during training; the CLIP
vision encoder remains frozen.

\section{Experiments}
\label{sec:experiments}

\subsection{Datasets and Evaluation Protocol}
\label{subsec:dataset}

NYU Depth V2~\cite{silberman2012indoor} is our primary benchmark. Models are
trained on the NYU training split and evaluated on the official test split using
valid ground-truth pixels in $[0.001,10.0]$ m. The reproduced SPACE-CLIP
baseline, SPACE-CLIPv2, and all ablations use identical preprocessing,
checkpoint selection, and evaluation code, with absolute predictions evaluated
without median scaling.

We additionally evaluate the same NYU-trained checkpoints zero-shot on
iBims-1~\cite{koch2019evaluation} and DIODE~\cite{vasiljevic2019diode}. For
iBims-1, we use all 100 official core images without target-domain adaptation,
calibration, direct-metric scale fitting, cropping, or TTA, and report both
standard depth and boundary/planarity measures. For DIODE, a single evaluation
covers all 771 official validation images (325 indoor and 446 outdoor), and we
report the indoor subset as a macro-average over per-image metrics. DIODE
evaluation likewise uses no target-domain adaptation, median scaling,
scale/shift fitting, calibration, cropping, or TTA.

\subsection{Implementation Details}
\label{subsec:implementation_details}

SPACE-CLIPv2 is implemented in PyTorch~\cite{paszke2019pytorch} using a
pre-trained CLIP ViT-B/16 image encoder~\cite{radford2021learning}. The CLIP image encoder is kept frozen throughout training and
inference, and the CLIP text encoder and text prompts are not used. All
trainable parameters are placed in the depth decoder and its adaptation modules.

The decoder receives grouped CLIP patch-token maps from selected transformer
layers and predicts dense metric depth. No external depth teacher or oracle
depth is used in the main experiments. Single-scale predictions are evaluated
without horizontal-flip ensembling or test-time depth alignment.

For reference, the evaluated fixed-local configuration ($K=4$) has 103.99M
parameters, of which 18.19M are trainable. On an NVIDIA RTX 5090 with batch-1
BF16 inference at $224\times224$ CLIP input resolution, it runs at
$11.77\pm0.08$ ms/image (84.995 images/s), with 725.21 MiB peak allocated
memory and an estimated 172.33 GFLOPs; an eight-epoch training run takes
approximately 2.32 h.

\subsection{Checkpoint Selection and Evaluation}
\label{subsec:checkpoint_selection}

During training, checkpoints are selected according to RMSE on the configured
checkpoint-selection split. The selected checkpoint is then evaluated under the
NYU test protocol to obtain the reported metrics. We use the same
checkpoint-selection rule for the reproduced SPACE-CLIP baseline, SPACE-CLIPv2,
and all internal ablation variants.

We report standard monocular depth estimation metrics, including AbsRel, SqRel,
RMSE, RMSE-log, SILog, log10 error, and threshold accuracies
$\delta_1$, $\delta_2$, and $\delta_3$~\cite{eigen2014depth,koch2019evaluation}.
When available, we also report boundary
AbsRel to measure depth accuracy near local geometric discontinuities. For
iBims-1, we additionally use depth-boundary error accuracy and completeness
(DBE-Acc. and DBE-Comp.) together with the official planar flatness and
orientation measures~\cite{koch2019evaluation}.

\subsection{Compared Methods}
\label{subsec:compared_methods}

We compare SPACE-CLIPv2 with two types of CLIP-based baselines. First, we report
published numbers from prior CLIP-based monocular depth estimation
methods~\cite{zhang2022depthclip,miya2026pureclipdepth,son2024cabins,kim2024clip2depth}.
Second, we include a SPACE-CLIP~\cite{cho2026spaceclipspatialperceptionadaptive}
baseline evaluated under the same setting as
SPACE-CLIPv2. This matched comparison is the primary evidence for measuring the
effect of our local geometry module.

\subsection{Ablation Study}
\label{subsec:ablation_study}

The ablation study isolates which components account for the improvement over
the reproduced SPACE-CLIP baseline. We first compare a decoder without local
aggregation against variants that add fixed or learned-offset aggregation. We
then compare learned-offset aggregation with the proposed fixed local stencil.
Finally, we evaluate token-space detail variants to test whether shallow CLIP
token maps provide complementary local contrast near the final depth prediction
head. All ablation variants use the same NYU split and checkpoint-selection rule.
To test sensitivity to initialization, we additionally repeat the
fixed-versus-learned comparison across five randomly selected NYU seeds.

\section{Results}
\label{sec:results}

\subsection{Local Token Aggregation Ablation}
\label{subsec:local_aggregation_results}

Table~\ref{tab:core_ablation} isolates the main architectural component of
SPACE-CLIPv2. Removing local aggregation gives an AbsRel of 0.123 and an RMSE of
0.428. Adding local aggregation improves both metrics: the learned-offset
variant reduces RMSE to 0.420 and increases $\delta_1$ from 0.853 to 0.860,
while the offset-free variant further reduces RMSE to 0.417 and reaches the
highest $\delta_1$ of 0.863. Boundary AbsRel also improves from 0.122 without
local aggregation to 0.118 with local aggregation. These results indicate that
local token evidence is useful for dense depth prediction, and that a fixed
local sampling pattern is at least as effective as learning sampling offsets
under a frozen CLIP backbone.

\begin{table*}[t]
\centering
\footnotesize
\setlength{\tabcolsep}{3.0pt}
\caption{Local token aggregation ablation on NYU Depth V2. B-AbsRel denotes
boundary AbsRel. All rows use the same token-detail setting.}
\label{tab:core_ablation}
\begin{tabular}{l c r r r r}
\toprule
Variant & Aggregation & AbsRel $\downarrow$ & RMSE $\downarrow$ &
$\delta_1 \uparrow$ & B-AbsRel $\downarrow$ \\
\midrule
Without local aggregation & -- & 0.123 & 0.428 & 0.853 & 0.122 \\
Learned-offset aggregation & learned & 0.119 & 0.420 & 0.860 & 0.118 \\
\textbf{Offset-free aggregation} & \textbf{fixed} & 0.119 &
\textbf{0.417} & \textbf{0.863} & 0.118 \\
\bottomrule
\end{tabular}
\end{table*}

\paragraph{Multi-seed robustness.}
Across five randomly selected NYU seeds, fixed-local sampling outperforms the
learned-offset variant in all five runs. Fixed sampling achieves
$0.4221\pm0.0061$ RMSE and $0.1194\pm0.0013$ AbsRel, compared with
$0.4238\pm0.0067$ and $0.1201\pm0.0024$, respectively, for learned offsets.
Although the average margin is modest, its consistent direction indicates that
the fixed-stencil result is not driven by a single favorable initialization.

\subsection{Cross-Domain Local Geometry}
\label{subsec:cross_domain}

Table~\ref{tab:cross_domain} evaluates the NYU-trained checkpoints zero-shot on
iBims-1. SPACE-CLIPv2 reduces AbsRel from 0.3353 to 0.3296 and increases
$\delta_1$ from 0.4164 to 0.4434. More importantly, DBE accuracy and
completeness improve by 20.35\% and 40.70\%, respectively, while both planar
measures also improve. Paired image-level bootstrap
intervals~\cite{efron1979bootstrap} for the SPACE-CLIPv2 minus SPACE-CLIP
difference exclude zero for DBE accuracy
($[-1.539,-0.565]$ px) and completeness ($[-53.489,-27.427]$ px).

\begin{table}[t]
\centering
\footnotesize
\setlength{\tabcolsep}{3.0pt}
\caption{Zero-shot iBims-1 evaluation on all 100 official core images. Both
models are trained only on NYU. Lower is better except for $\delta_1$.}
\label{tab:cross_domain}
\begin{tabular}{lrrr}
\toprule
Metric & SPACE-CLIP & SPACE-CLIPv2 & Change \\
\midrule
AbsRel $\downarrow$ & 0.3353 & \textbf{0.3296} & $-1.70\%$ \\
$\delta_1 \uparrow$ & 0.4164 & \textbf{0.4434} & $+6.48\%$ \\
DBE-Acc. (px) $\downarrow$ & 5.124 & \textbf{4.081} & $-20.35\%$ \\
DBE-Comp. (px) $\downarrow$ & 98.743 & \textbf{58.560} & $-40.7\%$ \\
Flatness (cm)$^\dagger$ $\downarrow$ & 18.972 & \textbf{17.709} & $-6.66\%$ \\
Orientation ($^\circ$)$^\dagger$ $\downarrow$ & 45.330 & \textbf{41.418} & $-8.63\%$ \\
\bottomrule
\end{tabular}
\vspace{1mm}

\par\smallskip
\parbox{\linewidth}{%
    \scriptsize
    \noindent$^\dagger$\,The official planarity protocol uses per-plane
    ground-truth-based median alignment; these measures therefore provide
    complementary planar-geometry evidence rather than raw metric-depth accuracy.
}
\end{table}

On the indoor subset of the official DIODE validation archive (325 images),
SPACE-CLIPv2 also reduces macro-averaged AbsRel from 0.4405 to 0.4309 using the
same NYU-trained checkpoints. This result provides complementary cross-domain
evidence; not all individual DIODE metrics improve, so we do not interpret it as
uniform cross-domain superiority.

\subsection{Comparison with CLIP-Based Depth Estimation}
\label{subsec:clip_depth_comparison}

Table~\ref{tab:nyu_clip_based_depth} places SPACE-CLIPv2 among CLIP-based
monocular depth estimation methods on NYU Depth V2. Because prior methods may
use different training and inference protocols, the most direct comparison is
between SPACE-CLIP and SPACE-CLIPv2 under our reproduced setting. SPACE-CLIPv2
improves over SPACE-CLIP from 0.123 to 0.119 in AbsRel and from 0.435 to 0.417
in RMSE. It also reduces log10 error from 0.053 to 0.051 and improves
$\delta_1$ from 0.849 to 0.863.

\begin{table}[t]
\centering
\footnotesize
\setlength{\tabcolsep}{2.0pt}
\caption{CLIP-based monocular depth estimation on NYU Depth V2. Published
methods may use different protocols; SPACE-CLIP and SPACE-CLIPv2 use our
matched evaluation setting.}
\label{tab:nyu_clip_based_depth}
\begin{tabular}{l r r r r}
\toprule
Model & AbsRel $\downarrow$ & RMSE $\downarrow$ & log10 $\downarrow$ &
$\delta_1 \uparrow$ \\
\midrule
DepthCLIP~\cite{zhang2022depthclip} & 0.388 & 1.167 & 0.156 & 0.394 \\
PureCLIP-Depth~\cite{miya2026pureclipdepth} & 0.201 & 0.670 & 0.084 & 0.671 \\
CaBins~\cite{son2024cabins} & 0.120 & 0.401 & 0.050 & 0.866 \\
CLIP2Depth~\cite{kim2024clip2depth} & \textbf{0.100} & \textbf{0.379} &
\textbf{0.042} & \textbf{0.900} \\
\midrule
SPACE-CLIP~\cite{cho2026spaceclipspatialperceptionadaptive} &
0.123 & 0.435 & 0.053 & 0.849 \\
\textbf{SPACE-CLIPv2} & \textbf{0.119} & \textbf{0.417} &
\textbf{0.051} & \textbf{0.863} \\
\bottomrule
\end{tabular}
\end{table}

\begin{figure}[t]
    \centering
    \includegraphics[width=0.95\linewidth]{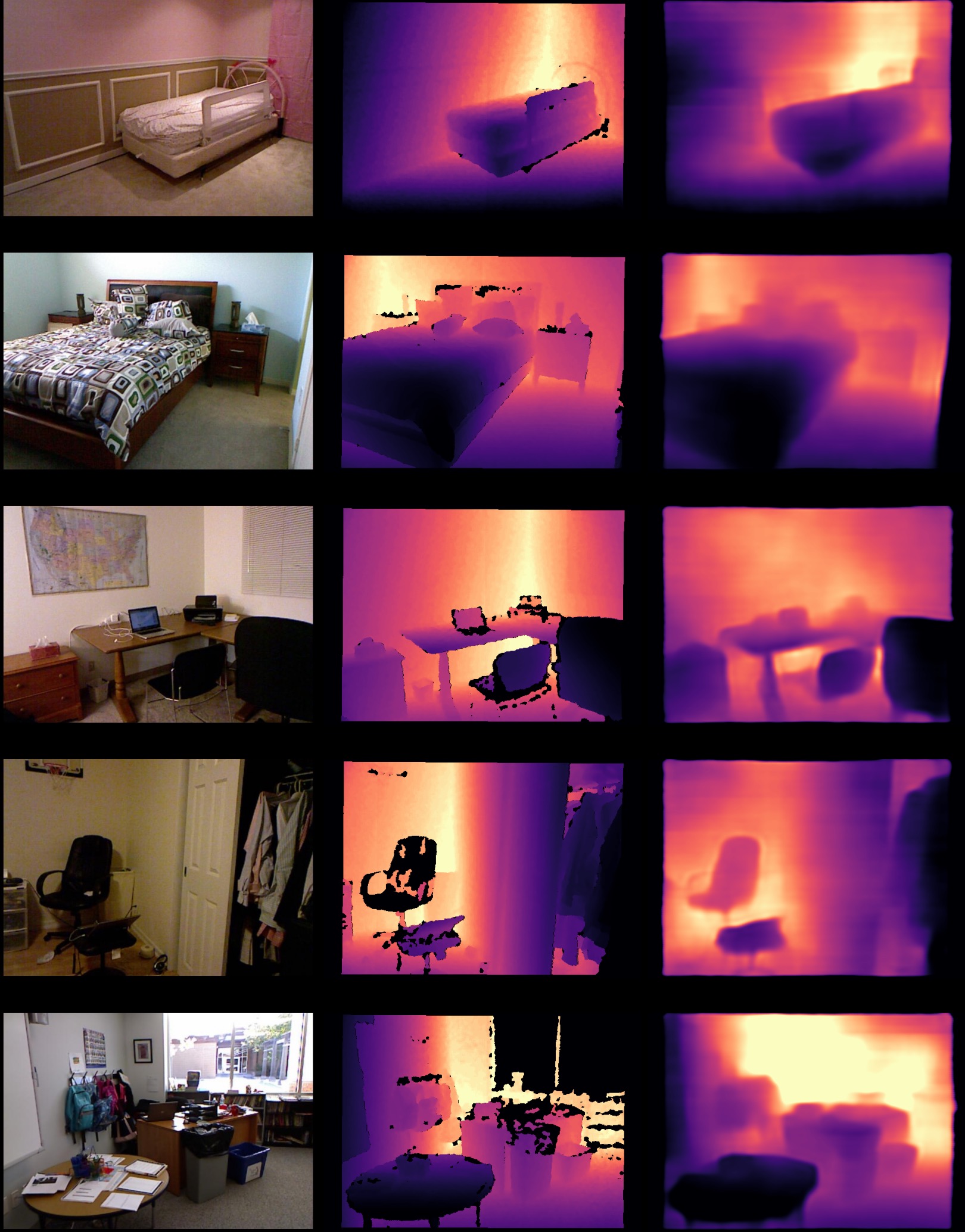}
    \caption{Qualitative NYU Depth V2 examples. Each row shows RGB input,
    ground-truth depth, and SPACE-CLIPv2 prediction. Predictions preserve the
    main room layout and large planar structures, while remaining errors mainly
    occur near thin objects, boundaries, and invalid ground-truth regions.}
    \label{fig:nyu_qualitative}
\end{figure}

\subsection{Token-Space High-Pass Detail}
\label{subsec:highpass_results}

The token-space high-pass branch provides complementary shallow detail. Before
local aggregation is introduced, it improves the reproduced decoder from
0.133 to 0.123 AbsRel and from 0.446 to 0.428 RMSE. Combining it with local
aggregation gives the strongest result, 0.119 AbsRel and 0.417 RMSE
(Table~\ref{tab:core_ablation}), so we treat it as a supporting rather than
central component.

\section{Conclusion}
\label{sec:conclusion}

We presented SPACE-CLIPv2, a decoder framework for recovering local geometry
from frozen CLIP representations through fixed local token aggregation, gated
residual injection, and complementary token-space detail refinement. Under
matched NYU Depth V2 evaluation, local aggregation improves both a decoder
without local aggregation and the reproduced SPACE-CLIP baseline, while
fixed-local sampling consistently outperforms learned offsets across five
randomly selected seeds.

Zero-shot iBims-1 evaluation further shows substantial improvements in depth
boundary errors together with improved planar-geometry measures, and indoor
DIODE provides complementary cross-domain AbsRel evidence. These results
suggest that a constrained locality prior is a practical decoder-side mechanism
for frozen CLIP features. The contribution lies in this depth-oriented
integration rather than in any individual aggregation primitive. SPACE-CLIPv2
retains a frozen, text-free CLIP backbone without an external depth teacher or
test-time alignment; broader cross-domain and downstream evaluation remain
directions for future work.

\section*{Acknowledgements}
This work was supported by Institute of Information \& communications Technology Planning \& Evaluation(IITP) under the Leading Generative AI Human Resources Development(IITP-2026-RS-2026-25545512) grant funded by the Korea government(MSIT) and by the Technology Innovation Program(Semiconductor Advanced Industry Technology Development Project)(RS-2026-25522959, Development of Multimodal On-sensor PIM-AI Semiconductor and Sensory Data-based Intelligent Motion Control Technology for Humanoid Robot Hands) funded By the Ministry of Trade Industry \& Energy(MOTIE, Korea).

\bibliographystyle{splncs04}
\bibliography{main}

\end{document}